\documentclass[conference]{IEEEtran}
\IEEEoverridecommandlockouts

\usepackage[T1]{fontenc}
\usepackage{cite}
\usepackage{amsmath}
\usepackage{amssymb}
\usepackage{amsfonts}
\usepackage{bm}
\usepackage{graphicx}
\usepackage{subcaption}
\usepackage{quantikz}
\usepackage{xcolor}
\usepackage{url}
\usepackage{physics}
\usepackage[utf8]{inputenc}
\usepackage[hidelinks]{hyperref}

\newcommand{\blfootnote}[1]{{%
  \renewcommand{\thefootnote}{}\footnotetext{#1}%
  \addtocounter{footnote}{-1}}}

\begin{document}

% --- TITLE & AUTHORS ---
\title{Implementations of Quantum and Classical Topology-Aligned Architectures for Molecular Property Prediction\\
}

\author{\IEEEauthorblockN{1\textsuperscript{st} James T. Pegg}
\IEEEauthorblockA{\textit{QunaSys Europe} \\
Copenhagen, Denmark \\
james@qunasys.com \\
\href{https://orcid.org/0000-0002-6743-8651}{0000-0002-6743-8651}}
\and
\IEEEauthorblockN{2\textsuperscript{nd} Hubert Okadome Valencia}
\IEEEauthorblockA{\textit{QunaSys Inc.} \\
Tokyo, Japan \\
hubert@qunasys.com \\
\href{https://orcid.org/0000-0003-0955-3886}{0000-0003-0955-3886}}
\and
\IEEEauthorblockN{3\textsuperscript{rd} Ronin Wu}
\IEEEauthorblockA{\textit{QunaSys Europe} \\
Copenhagen, Denmark \\
ronin@qunasys.com \\
\href{https://orcid.org/0000-0002-3736-5391}{0000-0002-3736-5391}}
}

\maketitle
\blfootnote{\textcopyright~2026 IEEE. Personal use of this material is permitted. Permission from IEEE must be obtained for all other uses, in any current or future media, including reprinting/republishing this material for advertising or promotional purposes, creating new collective works, for resale or redistribution to servers or lists, or reuse of any copyrighted component of this work in other works. Accepted for publication in the 2026 IEEE International Conference on Quantum Computing and Engineering (QCE).}
\begin{abstract}
For low-data and resource-constrained regimes typical of quantum chemistry, parameter-efficient learning is a key objective. Here, we propose a topology-aligned inductive bias in which the model architecture mirrors the molecular bond graph: atoms map to a fixed register of computational units, and bonds determine which pairs interact through shared learnable parameters. This principle is instantiated in two architectures: a variational quantum circuit (Iso-QGNN) and a parameter-matched classical message-passing network (Iso-CGNN). The models are benchmarked on HOMO-LUMO and dipole moment binary classification tasks over the QM9 benchmark. With 64 trainable parameters, the implementations achieve test AUCs of approximately 0.89 (quantum) and 0.92 (classical) on the gap task, and close to 0.78 (both) on the dipole task. The models reach 90\% of asymptotic performance within about 300 training molecules and gradient norms remain stable throughout training. These results indicate that the topology-aligned inductive bias is the active ingredient driving parameter efficiency at QM9 scale, with implications for matched-baseline benchmarking in quantum machine learning.
\end{abstract}

\section{Introduction}

In the design of machine-learning interatomic potentials (MLIPs) and generative adversarial networks (GANs), graph-based methodologies have emerged as a primary tool for mapping molecular systems~\cite{batzner2022equivariant, WANG2024109673, decao2018molgan, thomas2025qca, mniszewski2021hybrid}. Within computational chemistry, the simulation of systems with intricate electronic structure (highly correlated, photoreactive, or excited-state systems) scales unfavourably; consequently, physicochemical property prediction for complex systems is often intractable from first principles~\cite{bartlett2007coupled, feynman1982simulating, cao2019quantum}. In cheminformatics, classical machine learning (ML) offers inference speeds orders of magnitude faster than ab initio calculation~\cite{butler2018machine, smith2017ani}. Here, graph neural networks (GNNs) have established themselves as the state-of-the-art; however, capturing subtle electronic correlations typically requires deep, highly parameterised architectures~\cite{fung2021benchmarking}. These in turn demand large training datasets that are frequently scarce or expensive.

Two distinct lines of work have responded to this data-and-parameters problem. The first, from the quantum machine learning (QML) community, encodes molecular information directly into the high-dimensional Hilbert space of a quantum processor; the appeal is a potentially favourable expressivity--parameter trade-off and improved data efficiency~\cite{havlivcek2019supervised, abbas2021power}. The second, from classical geometric deep learning, builds the symmetries and structure of the data into the model architecture itself, reducing the parameters needed to represent physically valid functions~\cite{bronstein2021geometric, satorras2021n, hoogeboom2022equivariant}. These lines aim at the same target: competitive predictive performance with fewer parameters and less data.

Matched-baseline benchmarking has emerged as a methodological focus in QML, motivated by the need to disentangle the contributions of these two design choices. Bowles et al.~\cite{bowles2024better} systematically compared twelve quantum models against classical baselines across 160 datasets and found that out-of-the-box classical methods generally outperformed their quantum counterparts on small-scale tasks, with the further observation that removing entanglement often left quantum-model performance unchanged. Subsequent benchmarking of quantum kernel methods has reached similar conclusions~\cite{egginger2025quantum, alvarezestevez2024benchmarking}. These results suggest that, in the noisy intermediate-scale quantum (NISQ) regime, the inductive bias of a model architecture often matters more than whether its substrate is quantum; performance claims for quantum models are most informative when the comparison controls for architectural design.

In the NISQ era, deploying effective quantum neural networks (QNNs) remains challenging~\cite{preskill2018quantum}. Earlier methods used quantum annealing for molecular similarity and electronic structure, but gate-based variational methods have become the dominant paradigm~\cite{hen2012solving, hernandez2017enhancing, streif2019solving, omalley2016scalable}. These methods generally rely on hardware-efficient ansätze (HEAs), where gates are arranged according to the connectivity of the processor rather than the structure of the problem~\cite{kandala2017hardware, bharti2022noisy}. Such generic circuits are agnostic to chemical topology and lack inductive bias~\cite{gard2020efficient}. To capture complex chemical features, HEAs must therefore be deepened, introducing a critical trade-off: deep variational circuits are notoriously susceptible to gate-error accumulation and barren plateaus, where gradients vanish exponentially with system size~\cite{mcclean2018barren, cerezo2021cost, fujii2022deep}.

In this work, we examine an alternative: a structure-preserving architecture in which the model topology directly mirrors the molecular graph. Atoms are mapped to a fixed register of computational units, and bonds determine which pairs interact through shared, learnable parameters. Topology-aware quantum graph models for QM9-class tasks have been explored before~\cite{ryu2023quantum}, but typically without parameter-matched classical controls. We instantiate the topology-aligned inductive bias in two architectures: a variational quantum circuit (Iso-QGNN), in which entangling operations are applied between qubits corresponding to bonded atoms, and a parameter-matched classical message-passing network (Iso-CGNN), in which bonded nodes exchange messages weighted by the same shared parameters. For the two implementations with identical parameter count and identical splits, the contribution of the inductive bias is relatively isolated from that of the quantum substrate.

Our contributions are threefold. First, we show that the topology-aligned inductive bias yields competitive performance on QM9 binary classification with only 64 trainable parameters in both implementations, reaching test AUCs of 0.89 (quantum) and 0.92 (classical) on the highest occupied--lowest unoccupied molecular orbital (HOMO-LUMO) gap task, well above a non-graph logistic-regression control evaluated on identical splits. Second, we show that both implementations are highly data-efficient, reaching 90\% of asymptotic performance within about 300 training molecules. Third, the matched comparison demonstrates that the inductive bias rather than the quantum substrate is the active ingredient driving parameter efficiency at QM9 scale, with the quantum and classical implementations performing comparably across all metrics tested. We discuss what this result implies for QML benchmarking methodology and for the regimes in which a quantum implementation may offer genuine advantages.

\section{Method}
\label{sec:method}

\subsection{Dataset and Graph Representation}
We use the QM9 benchmark~\cite{ramakrishnan2014quantum}, comprising approximately 134{,}000 small organic molecules with computed geometric and electronic properties. We address two binary classification tasks: the HOMO-LUMO energy gap ($\Delta\epsilon$) and the electric dipole moment ($\mu$). Both targets are scalars as tabulated in QM9: $\Delta\epsilon$ is an orbital-energy difference and $\mu$ is the magnitude of the dipole vector. This framing focuses the comparison on architectural efficiency rather than absolute regression accuracy.

Binary labels are produced by a median split. All experiments draw a fixed sample of 1{,}500 molecules from QM9 (seeded), and each continuous target value is assigned label 1 if it exceeds the median and 0 otherwise. To keep the labels free of any dependence on held-out data, the median threshold is computed on the training split only (recomputed per trial) and applied unchanged to the validation and test molecules; the training set is therefore exactly balanced by construction, while the held-out splits are close to balanced. The two architectures and the non-graph control receive byte-identical labels within each trial, so the comparison remains matched.

Each molecule is represented as a graph $G = (V, E)$, where nodes $v_i \in V$ correspond to heavy atoms (C, N, O, F) and edges $e_{ij} \in E$ represent chemical bonds. Hydrogen atoms are treated implicitly. To ensure a constant register width across batches, we fix the maximum heavy-atom count at $N_{\max} = 9$; molecules exceeding this are excluded, and smaller molecules are padded with non-interacting ghost nodes. In the quantum implementation these are qubits initialised in $|0\rangle$ to which no entangling or rotation gates are applied; in the classical implementation they are nodes whose state is held at zero and that participate in no message passing. This padding strategy ensures consistent tensor dimensions across molecules. As ghost nodes are non-interacting, they contribute a constant per-qubit readout ($\langle Z \rangle = +1$ in the quantum model, and a learned near-zero constant in the classical model); the number of such entries reflects heavy-atom count, a size signal that is applied identically to both architectures and is likewise available to the composition-based control, so it confers no quantum-versus-classical advantage.

Performance is evaluated with the area under the receiver operating characteristic curve (AUC), computed by integrating the true positive rate against the false positive rate across all decision thresholds of the raw classifier logits.

\subsection{Topology-Aligned Architectures}
We instantiate the same topology-aligned inductive bias in two architectures. The models share an identical input representation, parameter-sharing scheme, and classical readout head, and differ only in the primitive used to propagate information between bonded atoms. This construction makes the matched comparison structural rather than coincidental: differences in performance reflect the choice of substrate, not the choice of architecture.

\subsubsection{Iso-QGNN: quantum implementation}
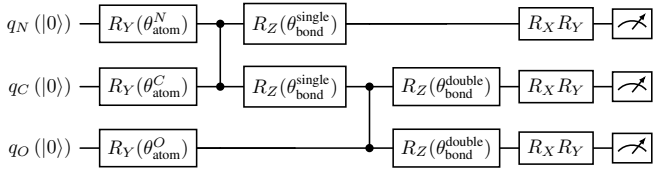
\begin{figure}[tb]
    \centering
    \resizebox{0.48\textwidth}{!}{%
    \begin{quantikz}[row sep=0.4cm, column sep=0.35cm]
        \lstick{$q_N\,(\ket{0})$} & \gate{R_Y(\theta_{\text{atom}}^{N})} & \ctrl{1}   & \gate{R_Z(\theta_{\text{bond}}^{\text{single}})} & \qw        & \qw                                              & \gate{R_X R_Y} & \meter{} \\
        \lstick{$q_C\,(\ket{0})$} & \gate{R_Y(\theta_{\text{atom}}^{C})} & \control{} & \gate{R_Z(\theta_{\text{bond}}^{\text{single}})} & \ctrl{1}   & \gate{R_Z(\theta_{\text{bond}}^{\text{double}})} & \gate{R_X R_Y} & \meter{} \\
        \lstick{$q_O\,(\ket{0})$} & \gate{R_Y(\theta_{\text{atom}}^{O})} & \qw        & \qw                                              & \control{} & \gate{R_Z(\theta_{\text{bond}}^{\text{double}})} & \gate{R_X R_Y} & \meter{}
    \end{quantikz}%
    }
    \caption{\textbf{Iso-QGNN circuit for a three-atom example (formamide O=CH--NH$_2$, with implicit H atoms).} Each atom is assigned a qubit initialised in $\ket{0}$ and rotated by $R_Y(\theta_{\text{atom}})$ with a per-atom-type parameter. Bonded atoms are then entangled with a $CZ$ gate followed by symmetric $R_Z(\theta_{\text{bond}})$ rotations with per-bond-type parameters (single and double bonds receive different parameters). A final $R_X, R_Y$ rotation layer is applied, and each qubit's magnetisation $\langle Z_i \rangle$ is measured. Padding qubits are omitted for clarity.}
    \label{fig:circuit}
\end{figure}

The quantum architecture (Fig.~\ref{fig:circuit}) proceeds in three stages.

\textit{Atom embedding.} For each atom type $A \in \{\text{C, N, O, F}\}$, a trainable parameter $\theta_{\text{atom}}^{(A)}$ is assigned. The initial state is prepared by applying $R_Y(\theta_{\text{atom}}^{(A)})$ to the qubit $q_i$ corresponding to each non-padding atom. These four parameters are initialised uniformly in $[0, 2\pi]$ to break symmetry at the start of training.

\textit{Isomorphic entanglement.} For every bond type $T \in \{\text{single, double, triple, aromatic}\}$, a controlled-phase gate ($CZ$) followed by symmetric local rotations $R_Z(\theta_{\text{bond}}^{(T)})$ is applied between the qubits of bonded atoms. This composition realises a parameterised diagonal interaction whose strength is shared across all bonds of a given type. Restricting the entangling primitive to a diagonal $CZ$ family and the readout to the $Z$ basis keeps gate depth and measurement overhead minimal at near-term scale, at the cost of limiting expressivity to diagonal correlations~\cite{kandala2017hardware}. It introduces a bond-type-dependent phase ($R_Z$) and an entanglement-inducing $CZ$, approximating but not exactly equivalent to a symmetric $R_{ZZ}$ interaction.

\textit{Variational rotations.} A layer of single-qubit rotations $R_X(\phi^{(1)}_q), R_Y(\phi^{(2)}_q)$ is applied to each non-padding qubit, allowing quantum interference and state mixing analogous to a node-update step in classical message passing.

The molecular representation is obtained by measuring the local magnetisation $\langle Z_i \rangle$ at each qubit. The resulting feature vector $\mathbf{x} \in \mathbb{R}^{N_{\max}}$ is passed to a lightweight classical head (LayerNorm followed by a linear classifier) producing the classification logits.

\subsubsection{Iso-CGNN: classical implementation}
The classical architecture mirrors Iso-QGNN element-by-element, replacing each quantum primitive with its closest classical analogue (Table~\ref{tab:mapping}). Each atom type carries a trainable scalar $\theta_{\text{atom}}^{(A)}$; the initial node state is $h_i = \cos(\theta_{\text{atom}}^{(A)})$, an analogue of the $\langle Z \rangle$ expectation produced by an $R_Y$ rotation on $|0\rangle$. For each bond type $T$ a trainable scalar weight $\theta_{\text{bond}}^{(T)}$ is shared across all bonds of that type. A message-passing update accumulates contributions from bonded neighbours weighted by these shared bond parameters: $h_i \leftarrow h_i + \sum_{j \in \mathcal{N}(i)} \theta_{\text{bond}}^{(T_{ij})}\, h_j$. A node-wise transform $h_i \leftarrow \tanh(s_i\, h_i + b_i)$ then plays the role of the variational rotation, with $\tanh$ keeping each node state bounded in $[-1, 1]$ as a $Z$-basis expectation would be. The same LayerNorm and linear classifier head produces the logits.

\begin{table}[t]
    \centering
    \caption{\textbf{Structural mapping between Iso-QGNN and Iso-CGNN.} Each row pairs a quantum primitive with its classical analogue at matched parameter count.}
    \label{tab:mapping}
    \setlength{\tabcolsep}{3.8pt}
    \begin{tabular}{ll}
        \hline
        \textbf{Iso-QGNN primitive} & \textbf{Iso-CGNN analogue} \\
        \hline
        $R_Y(\theta_{\text{atom}})$ on $|0\rangle$ & $\cos(\theta_{\text{atom}})$ initial state \\
        $CZ$ + $R_Z(\theta_{\text{bond}})$ on bond & shared bond-weighted message \\
        $R_X(\phi^{(1)}), R_Y(\phi^{(2)})$ rotations & $\tanh(s\,h + b)$ node update \\
        $\langle Z_i \rangle$ measurement & scalar node state ($\tanh$-bounded) \\
        LayerNorm + Linear classifier & LayerNorm + Linear classifier \\
        \hline
    \end{tabular}
\end{table}

\subsubsection{Parameter accounting}
The architectures use a single layer ($L=1$) and the parameter accounting is identical (Table~\ref{tab:parameters}). The atom and bond embeddings each contribute four shared parameters; the per-node update contributes two parameters per qubit (or node), giving $2 \times 9 = 18$. The classical readout head contributes 18 LayerNorm parameters ($\gamma$ and $\beta$) and 20 linear-classifier parameters (weight matrix and bias), for a total of 64 trainable parameters per architecture. This is the matched capacity used throughout the comparison.

\begin{table}[t]
    \centering
    \caption{\textbf{Parameter accounting.} The Iso-QGNN and Iso-CGNN models share the same breakdown at a single layer ($L=1$).}
    \label{tab:parameters}
    \begin{tabular}{lcc}
        \hline
        \textbf{Component} & \textbf{Symbol} & \textbf{Count} \\
        \hline
        Atom embedding (one per atom type) & $\theta_{\text{atom}}$ & 4 \\
        Bond parameter (one per bond type) & $\theta_{\text{bond}}$ & 4 \\
        Per-node update ($N_{\max} \times 2$) & $\phi$ or $s, b$ & 18 \\
        LayerNorm ($\gamma, \beta$) & --- & 18 \\
        Linear classifier (weight + bias) & $W, b$ & 20 \\
        \hline
        \textbf{Total} & & \textbf{64} \\
        \hline
    \end{tabular}
\end{table}

\subsection{Training Protocol}
The architectures are trained with the Adam optimiser ($lr = 0.01$) and cross-entropy loss. Parameters are updated after every training molecule (batch size one), and each model is trained for 200 epochs, by which point the trajectories have converged. To mitigate cold-start gradient issues in the quantum circuit, the bond-entanglement and variational parameters are initialised from $\mathcal{N}(0, 0.01)$, ensuring the circuit begins near the identity; the same scheme is applied to the corresponding parameters of Iso-CGNN for consistency. Atom embeddings are uniformly randomised in $[0, 2\pi]$ for both models.

For the matched-capacity comparison, we use ten independent trials with seeds $\{42, 43, \ldots, 51\}$; the sample-complexity sweep uses five trials per training size, reflecting its larger compute footprint (19 training sizes for each task and architecture). The training/validation/test split (80/10/10) is generated once per trial from a seeded random generator, and both architectures are trained on byte-identical splits within each trial. Before each model is constructed, all PyTorch, NumPy, and Python random number generators (RNGs) are reset to a common seed so that parameter initialisation draws from a matched RNG state. All quantum experiments are performed via state-vector simulation using PennyLane~\cite{bergholm2018pennylane}; no hardware noise model is included. All reported metrics are averaged over these ten seeds. The variational quantum models show sensitivity to initialisation, most visibly on the gap task where individual seeds can differ by up to $\sim 0.15$ AUC and shorter training runs occasionally leave a seed under-converged; we therefore report converged 200-epoch values with cross-trial standard deviations, and the released code pins exact dependency versions for reproducibility.

\section{Results and Discussion}

\subsection{Predictive Performance at Matched Capacity}
At identical parameter count (64) and on identical splits, both architectures achieve competitive performance on the QM9 binary classification tasks (Table~\ref{tab:performance}). On the HOMO-LUMO gap task, Iso-CGNN reaches a test AUC of 0.916 and Iso-QGNN reaches 0.887; on the dipole moment task, the two are level within noise at 0.773 (Iso-QGNN) and 0.786 (Iso-CGNN). Across the two tasks the implementations reach test AUCs from roughly 0.77 to 0.92 with 64 trainable parameters. The cross-trial standard deviations ($\pm 0.022$ to $\pm 0.042$) indicate that single runs of either architecture should be expected to land within roughly $\pm 0.04$ AUC of the reported mean.

To verify that the reported AUCs reflect learned structure rather than easy task geometry, we include a non-graph control: an $\ell_2$-regularised logistic regression on heavy-atom composition (the per-molecule counts of C, N, O, F; five trainable parameters including the bias), fit to convergence and evaluated with the same test AUC and ten-seed protocol as the graph models. The quoted values are mean test AUCs, not a saturation point: the control reaches only 0.659 (gap) and 0.698 (dipole), well below both topology-aligned models on both tasks. The median-split tasks are therefore not solvable from atomic composition alone, and the margin over this baseline is attributable to the information the architectures extract from the bond structure. On the HOMO-LUMO gap this margin reaches 0.23--0.26 AUC, several times the cross-trial standard deviation; whereas, on the dipole moment it narrows to roughly 0.08, exceeding the 0.022--0.042 spread by a smaller factor. A systematic study of richer hand-crafted molecular descriptors as non-graph baselines is the subject of ongoing work.

\begin{table}[t]
    \centering
    \caption{\textbf{Test performance at matched capacity.} Mean $\pm$ cross-trial standard deviation over ten trials on identical splits, trained to convergence (200 epochs). LogReg is a non-graph control: logistic regression on heavy-atom composition (atom-type counts). Bold marks the leading model on the gap task; the dipole task is a tie within noise.}
    \label{tab:performance}
    \setlength{\tabcolsep}{3pt}
    \begin{tabular}{llccc}
        \hline
        \textbf{Model} & \textbf{Property} & \textbf{Params} & \textbf{Test AUC} & \textbf{Test Acc.} \\
        \hline
        LogReg    & Gap ($\Delta\epsilon$)  & 5  & $0.659 \pm 0.045$ & $0.627 \pm 0.040$ \\
        Iso-QGNN  & Gap ($\Delta\epsilon$)  & 64 & $0.887 \pm 0.042$ & $0.795 \pm 0.040$ \\
        Iso-CGNN  & Gap ($\Delta\epsilon$)  & 64 & $\mathbf{0.916 \pm 0.022}$ & $\mathbf{0.815 \pm 0.026}$ \\
        \hline
        LogReg    & Dipole ($\mu$)          & 5  & $0.698 \pm 0.033$ & $0.649 \pm 0.024$ \\
        Iso-QGNN  & Dipole ($\mu$)          & 64 & $0.773 \pm 0.033$ & $0.717 \pm 0.052$ \\
        Iso-CGNN  & Dipole ($\mu$)          & 64 & $0.786 \pm 0.038$ & $0.723 \pm 0.042$ \\
        \hline
    \end{tabular}
\end{table}

A consistent task-level pattern emerges across the architectures: predictive accuracy is higher on the gap than on the dipole. The HOMO-LUMO gap is an eigenvalue difference associated with the system Hamiltonian, and the topological structure of bonding correlates strongly with frontier-orbital delocalisation. The dipole moment, in contrast, reflects the full molecular charge distribution (the nuclear point charges together with the electronic density), and its prediction fundamentally requires resolving the vector sum of partial charges. The architectures encode purely two-dimensional graph connectivity, with no explicit spatial geometry; consequently, they are subject to the same geometric information bottleneck on the dipole task.

The head-to-head comparison varies by task. On the gap task, Iso-CGNN exceeds Iso-QGNN by 2.8 AUC points and 1.9 accuracy points; at roughly 0.6 combined cross-trial standard deviations this is a modest lead rather than a sharp separation at matched capacity, but it is consistent, as the classical implementation leads across the entire sample-complexity sweep (Fig.~\ref{fig:efficiency}). On the dipole task the two architectures are statistically indistinguishable: the mean-AUC difference (0.773 vs 0.786) is far below one combined cross-trial standard deviation, a clear tie within noise. Iso-QGNN exhibits higher cross-trial standard deviation than Iso-CGNN on the gap task ($\pm 0.042$ vs $\pm 0.022$), indicating greater sensitivity to initialisation in the quantum implementation there; on the dipole task the two spreads are comparable ($\pm 0.033$ vs $\pm 0.038$).

\begin{figure}[tb]
    \centering
    \includegraphics[width=0.48\textwidth]{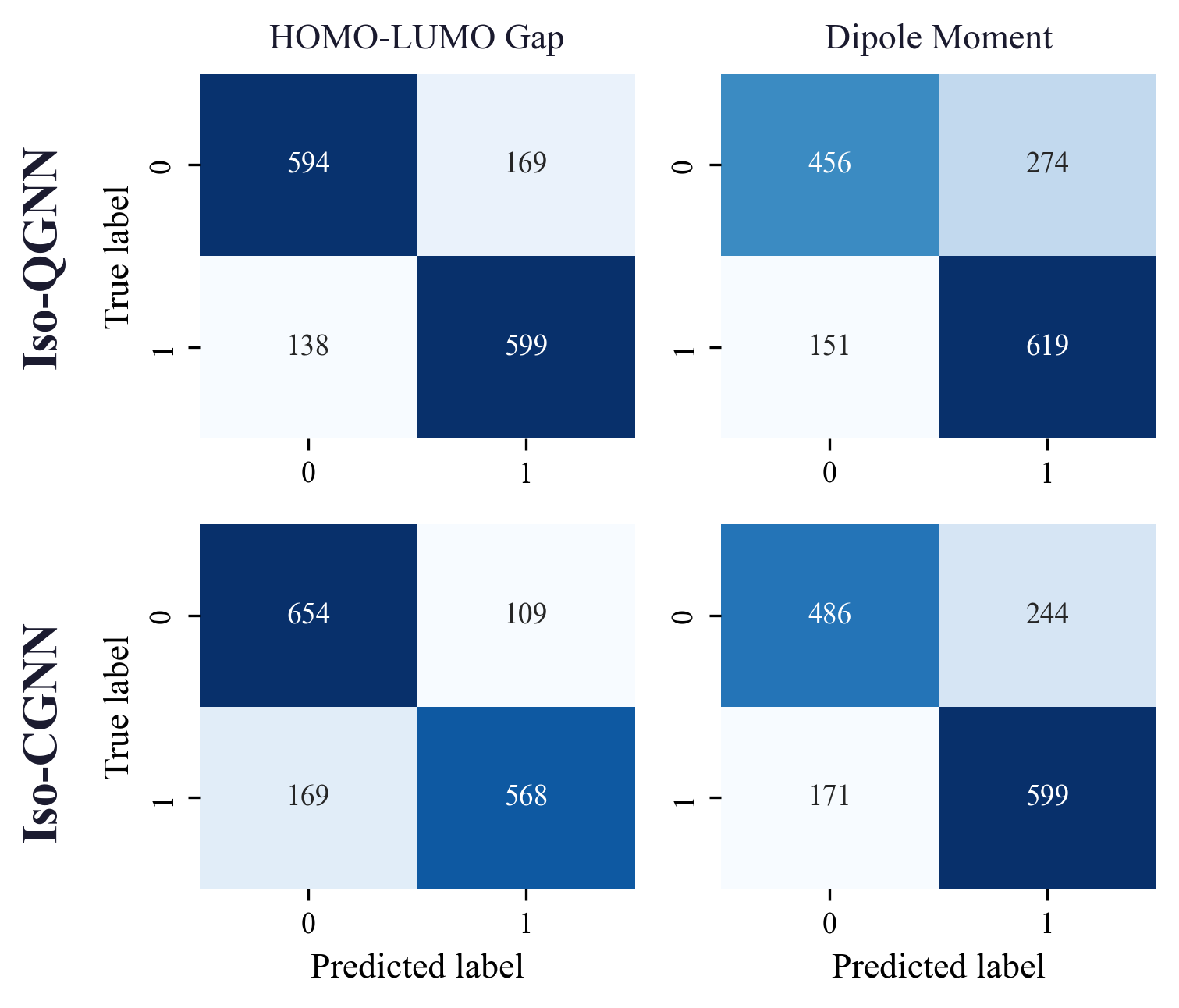}
    \caption{\textbf{Confusion matrices.} Aggregated test-set confusion over ten trials. Rows are the architectures (Iso-QGNN upper, Iso-CGNN lower); columns are the tasks (HOMO-LUMO gap left, dipole moment right). Diagonal dominance confirms both architectures learn discriminative features rather than class priors; Iso-CGNN's stronger diagonal on the gap task appears in the lower-left panel.}
    \label{fig:confusion}
\end{figure}

The aggregated confusion matrices (Fig.~\ref{fig:confusion}) confirm that both architectures concentrate their predictions along the diagonal and spread the residual errors across both off-diagonal cells, ruling out class-prior exploitation. Iso-CGNN's higher gap-task accuracy appears as a larger diagonal count in the lower-left panel (1222 vs 1193 correct of 1500).

\subsection{Sample Complexity}
\begin{figure}[tb]
    \centering
    \includegraphics[width=0.48\textwidth]{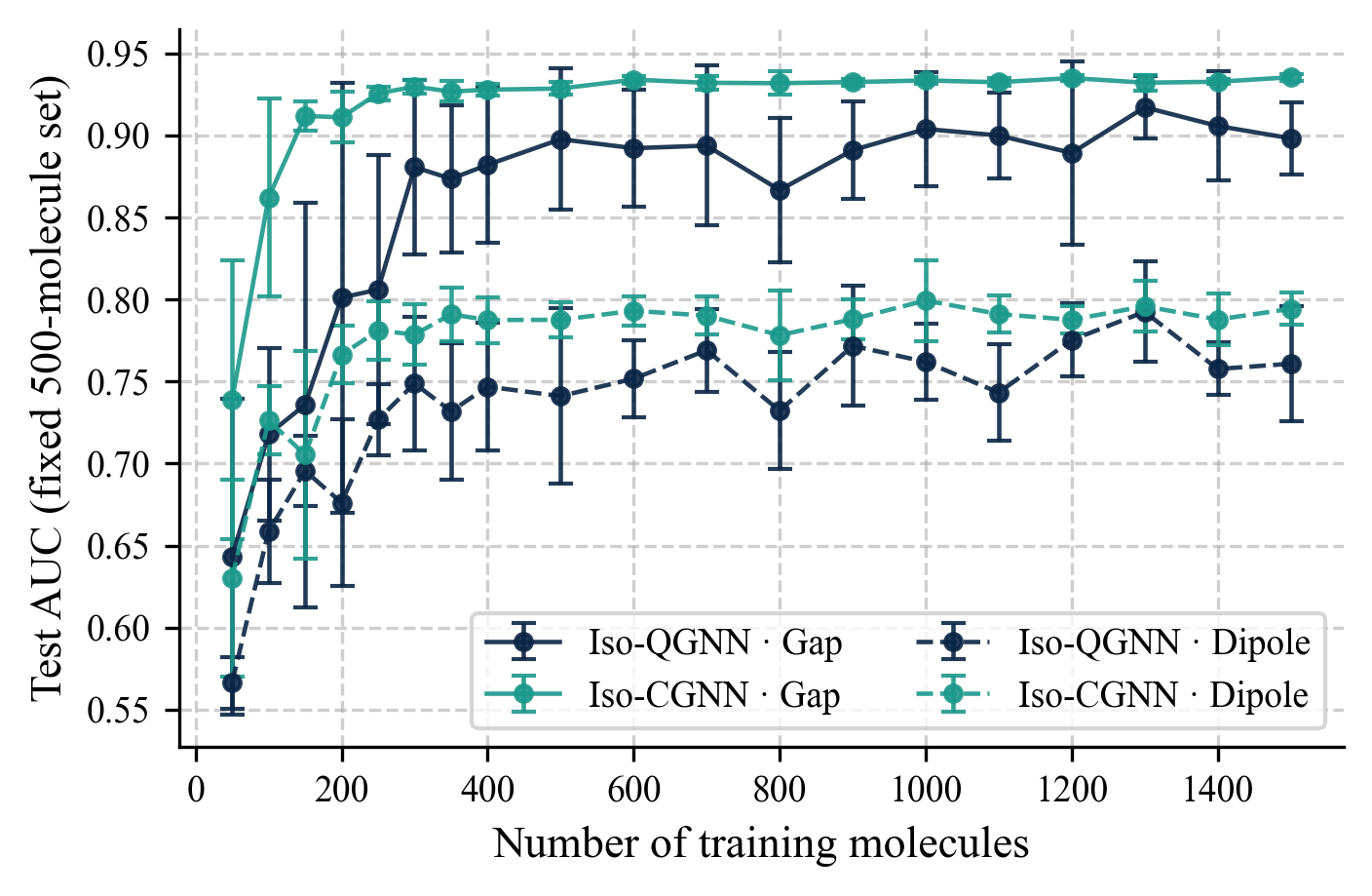}
    \caption{\textbf{Sample complexity.} Test AUC on a fixed 500-molecule held-out set as a function of training-set size. The architectures saturate within $N \le 300$ molecules. Iso-CGNN reaches saturation faster on both tasks and maintains a small but consistent lead on the gap task across all training sizes. Error bars are standard deviations across five trials. For each training-set size we report the test AUC at the epoch selected by validation AUC, applied identically to both architectures.}
    \label{fig:efficiency}
\end{figure}

The architectures are highly data-efficient (Fig.~\ref{fig:efficiency}). Defining the saturation point $N_{90\%}$ as the smallest training-set size at which the model first reaches 90\% of its own asymptotic peak AUC, Iso-QGNN reaches saturation at $N_{90\%} \approx 300$ on the gap task and $N_{90\%} \approx 250$ on the dipole task, while Iso-CGNN saturates earlier at $N_{90\%} \approx 100$ on both tasks. The implementations enter a strongly predictive regime within about 300 training molecules. For context, deep classical GNNs for molecular property prediction are typically trained on datasets several orders of magnitude larger~\cite{batzner2022equivariant}; we note, however, that those models address full regression rather than binary classification, so this contrast illustrates the difference in operating regime rather than a like-for-like sample-complexity comparison. This sample-complexity study uses an independent, larger molecule pool (3{,}000 molecules) with a fixed 500-molecule held-out test set, a shorter per-point training budget (40 epochs, for tractability across the full sweep), and validation-selected reporting (the epoch is chosen by validation AUC, identically for both architectures); its absolute AUCs are therefore not directly comparable to the converged, matched-capacity values in Table~\ref{tab:performance}, though the qualitative ordering between architectures is preserved.

The gap between the two architectures persists across the full sweep on the gap task: Iso-CGNN leads by roughly 0.10--0.18 AUC at the smallest training sizes (N = 50 to 200) and narrows to a few hundredths of AUC at the largest sizes. On the dipole task the curves nearly overlap, with Iso-CGNN marginally ahead at the smallest sizes ($\lesssim 0.09$ AUC, within or modestly above the combined cross-trial standard deviation, dipping to a near-tie at $N = 150$) before the two converge. The two architectures are strongly data-efficient at the hundreds-of-molecules scale; on the gap task Iso-CGNN saturates roughly threefold sooner ($N_{90\%} \approx 100$ vs $\approx 300$) and leads through the small-$N$ regime, whereas on the dipole task the two are closely matched.

\subsection{Gradient Stability}
Variational quantum circuits are notoriously susceptible to barren plateaus, where parameter gradients vanish exponentially with system size~\cite{mcclean2018barren}. The reduced parameter manifold induced by topology-aligned sharing might be expected to mitigate this; the matched comparison lets us test whether any such mitigation is quantum-specific.

\begin{figure}[tb]
    \centering
    \includegraphics[width=0.48\textwidth]{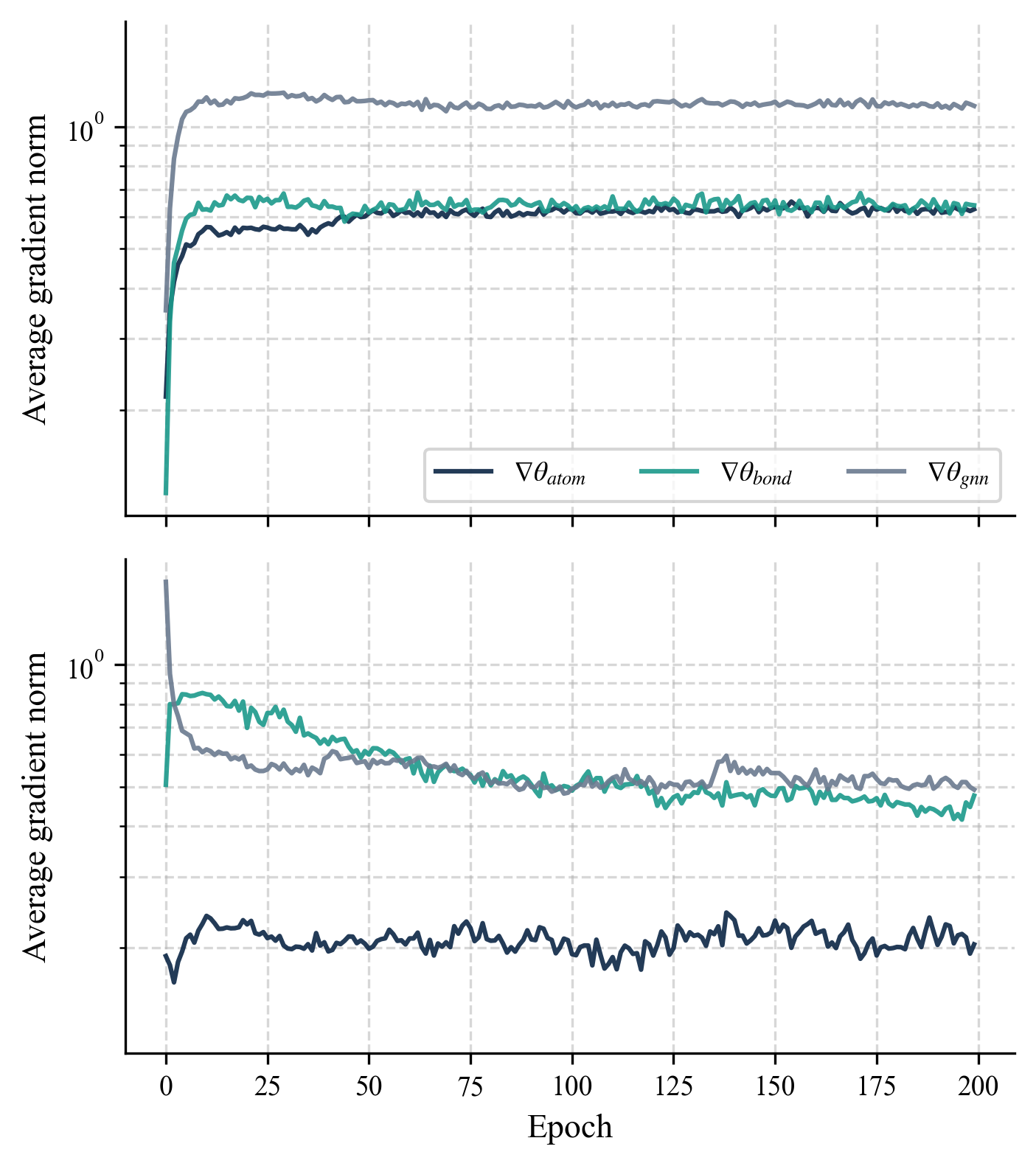}
    \caption{\textbf{Gradient flow stability.} Mean gradient norms (log scale) per parameter group across training on the gap task, for Iso-QGNN (upper graph) and Iso-CGNN (lower graph). Both maintain non-vanishing gradients (of order $10^{-1}$ to $\sim\!10^{0}$); the quantum implementation (upper) climbs from a small initialisation, while the classical one (lower) decays from a larger value. Stability holds at QM9 scale ($N \le 9$ heavy atoms); theoretical guarantees at larger system sizes remain open.}
    \label{fig:gradients}
\end{figure}

The architectures exhibit non-vanishing gradients throughout training (Fig.~\ref{fig:gradients}). For Iso-QGNN, gradient norms across all parameter groups remain of order $10^{-1}$ to $\sim\!10^0$ and stabilise within a few epochs. For Iso-CGNN, gradients of comparable magnitude remain stable across training, though with qualitatively different dynamics: where the quantum gradients climb from initialisation to a plateau, the classical gradients decay from a higher initial value to a comparable steady state. Neither architecture exhibits vanishing gradients at QM9 scale.

In both substrates, the shared property is therefore stable training under topology-aligned parameter sharing. We caution against extrapolating: barren-plateau scaling is an asymptotic property and the present work covers only $N \le 9$ qubits/nodes. Whether either architecture continues to benefit from stable gradients at larger system sizes is an open question.

\subsection{Bond-Parameter Analysis}
\begin{figure}[tb]
    \centering
    \includegraphics[width=0.48\textwidth]{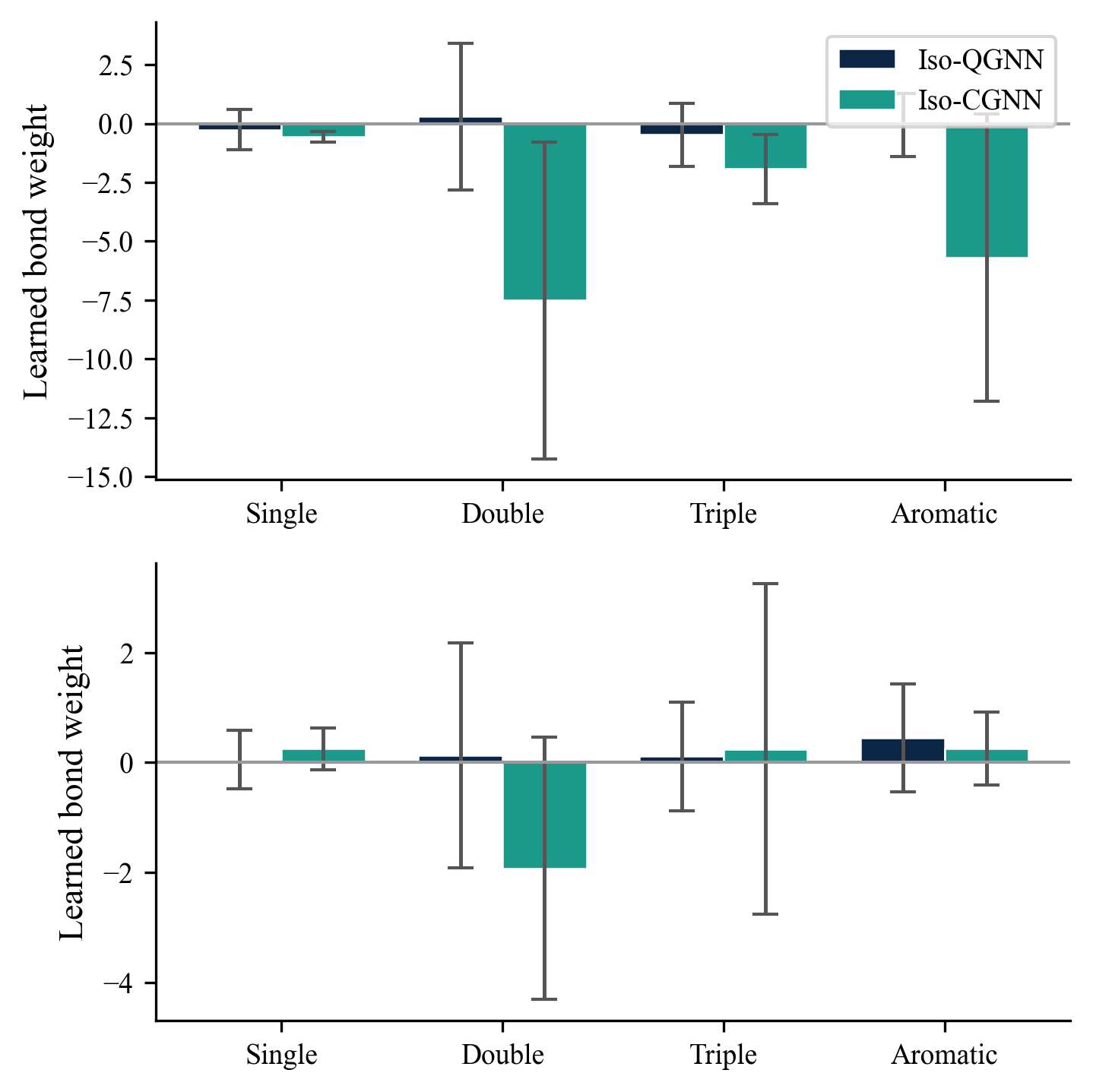}
    \caption{\textbf{Learned bond parameters.} Mean and standard deviation of the bond-type weights across ten trials, for both architectures on the HOMO-LUMO gap (upper graph) and dipole moment (lower graph) tasks. The large error bars indicate that neither architecture converges to a stable, individually interpretable bond hierarchy across seeds at QM9 scale; however, the underlying weight distributions remain task- and substrate-dependent (see text).}
    \label{fig:bonds}
\end{figure}

Earlier reports of single-trial runs of Iso-QGNN suggested a clean, task-dependent ranking of bond-parameter magnitudes that aligned with chemical intuition, with double, triple, and aromatic bonds receiving systematically larger entanglement strength than single bonds on the gap task. We do not find that this hierarchy is robust across seeds (Fig.~\ref{fig:bonds}). For both Iso-QGNN and Iso-CGNN, mean bond-type weights across ten seeds carry cross-trial standard deviations comparable to or larger than the means themselves on most bond types.

That said, the data carries a robust task-dependent signal even when the per-bond ranking does not stabilise. The bond-weight distributions differ between the gap and dipole tasks for both architectures, and they differ between the two architectures on the same task. The optimisation is therefore responding to chemical task structure: the model is not converging to noise, and the path through parameter space depends on which physical property is being predicted. What does not survive across seeds is the identification of any single bond type as uniquely important; the chemical signal is encoded in the joint distribution of all parameters together rather than in isolated bond weights.

This pattern suggests an under-explored direction. Even though post-training values of individual bond parameters are not directly interpretable, chemically informed initialisation (for example, scaling initial bond parameters by published bond enthalpies or related descriptors) could provide a more reliable starting point and yield more stable optimisation trajectories. Such priors would leverage the topology-aligned inductive bias more aggressively than the random initialisation used here, and could form a natural bridge between the architectural framework presented in this work and chemistry-aware constraint or correction schemes for downstream applications. We do not test this here, but flag it as a concrete future direction.

\subsection{Comparison}
\label{sec:comparison}
The matched-baseline comparison supports three substantive claims and clarifies one over-claim from the prior literature.

\textit{The inductive bias is the active ingredient.} Across the metrics tested (peak performance, data efficiency, gradient stability) both implementations of the topology-aligned architecture exhibit the same qualitative behaviour, with only modest task-dependent differences (Iso-CGNN saturates sooner on the gap task). The 64-parameter compactness, the data efficiency at the hundreds-of-molecules scale, and the stable gradient dynamics are properties of the topology-aligned parameter-sharing scheme, not of the quantum substrate.

\textit{Comparative claims in QML require matched controls.} Recent benchmarking work~\cite{bowles2024better, egginger2025quantum, alvarezestevez2024benchmarking} has shown that parameter-matched classical baselines often close apparent quantum advantages on small-scale tasks. Topical predecessors that report quantum advantage on QM9-class tasks~\cite{ryu2023quantum} did not include a fully matched classical control of the kind run here; the present results indicate that such controls are necessary methodology rather than optional due diligence.

\textit{Resource cost at near-term scale.} Iso-QGNN requires $N_{\max} = 9$ qubits with circuit depth scaling as the maximum graph degree (a small constant for organic molecules), and a per-layer gate count of approximately $3N + 3|E|$ (one $R_Y$ atom-encoding rotation plus two variational rotations per atom, and a $CZ$ with two $R_Z$ rotations per bond), dominated by operations on bonds. For QM9 at $L=1$ this puts the circuit comfortably within the coherence-time and gate-fidelity budget of current trapped-ion devices~\cite{bruzewicz2019trapped}. The shared parameter count is constant in $N_{\max}$ for both atom-type and bond-type embeddings; only the per-node update scales with $N_{\max}$, and linearly so.

\textit{Where the substrate may still matter.} On the gap task the classical implementation holds a small consistent lead; on the dipole task the two are statistically indistinguishable. With only ten trials and small effect sizes we cannot make strong claims about which substrate is preferable in general. More importantly, several regimes in which a quantum substrate has a theoretically motivated advantage are simply not exercised by the QM9 binary classification setting. Theoretical arguments for quantum utility in learning rest on feature maps that are hard to evaluate classically~\cite{havlivcek2019supervised} or on favourable effective-dimension and trainability properties of quantum models~\cite{abbas2021power}; neither mechanism is stressed by a nine-qubit circuit restricted to a diagonal entangling family with $Z$-basis readout, which was chosen deliberately for near-term feasibility. Concretely, we expect the substrate question to become non-trivial when (i) the target property depends on strongly correlated electronic structure whose classical surrogate models are themselves expensive~\cite{cao2019quantum}, (ii) molecule sizes grow beyond the reach of exact classical simulation of the corresponding circuits, or (iii) the readout requires non-diagonal observables that entangle the register. Whether topology-aligned quantum architectures retain or extend their parity with classical counterparts in those regimes is a natural direction for future work, and one that the present matched-comparison framework is well-suited to address: the classical control construction of Sec.~\ref{sec:method} carries over unchanged, while the quantum implementation gains access to the mechanisms above.

\subsection{Scaling Beyond QM9}
The QM9 benchmark restricts molecules to $N_{\max} = 9$ heavy atoms, but the architectural primitives in both implementations are constant or linear in molecule size, suggesting that resource cost grows gracefully with $N$. We summarise the scaling here, deferring empirical validation at larger $N$ to future work.

The trainable parameter count is dominated by the atom-type and bond-type embeddings (4 each in our setting) plus the per-node update; both atom-type and bond-type vocabularies are constant in $N$, and only the per-node update scales linearly. For Iso-QGNN, the qubit count scales as $O(N_{\max})$ by construction, and the per-layer gate count scales as $O(N_{\max} + |E|)$, with single-qubit rotations on each non-padding qubit and two-qubit entanglers on each bond. Iso-CGNN inference cost scales the same way: one update per node and one weighted message per edge.

The circuit depth scales as $O(d_{\max})$, where $d_{\max}$ is the maximum graph degree. Bonds incident on the same atom must be applied sequentially, but bonds at different atoms can parallelise. For organic molecules, $d_{\max}$ is bounded by carbon valence; this depth bound is independent of total molecule size. Concretely, a QM9 molecule with $N = 8$ and $|E| \approx 9$ requires approximately 51 gates per layer; a drug-like molecule at $N = 30$ and $|E| \approx 32$ requires approximately 186 gates per layer at the same depth budget.

This analysis does not predict whether either architecture's predictive accuracy is preserved at larger $N$. As graph diameter grows, a single layer ($L=1$) may no longer propagate information across the molecule, requiring additional layers and a corresponding increase in depth and parameter count. The matched-comparison framework introduced here translates directly to that regime, but empirical evaluation at $N > 9$ requires datasets covering larger molecules and is left for future work.

\section{Conclusion}

We presented a topology-aligned inductive bias for parameter-efficient molecular property prediction in which the model architecture mirrors the molecular bond graph: atoms map to a fixed register of computational units, and bonds determine which pairs interact through shared, learnable parameters. This principle was instantiated in two architectures: a variational quantum circuit (Iso-QGNN) and a parameter-matched classical message-passing network (Iso-CGNN). The models were benchmarked on QM9 binary classification of the HOMO-LUMO gap and dipole moment.

The matched comparison establishes three findings. With only 64 trainable parameters, both implementations reach test AUCs from roughly 0.77 to 0.92 across the two tasks, well above a five-parameter non-graph logistic-regression control on the same splits. The models reach 90\% of asymptotic performance within about 300 training molecules on the binary classification tasks studied here. And both maintain stable gradient flow throughout training at QM9 scale. These properties belong to the topology-aligned parameter-sharing scheme rather than to the quantum substrate: across all metrics tested, the quantum and classical implementations perform comparably, with Iso-CGNN holding a small consistent lead on the gap task and tying within noise on the dipole task.

We do not interpret this as a result against quantum machine learning. Rather, it clarifies what the source of the parameter efficiency is in this class of architecture, and it adds to a growing body of work~\cite{bowles2024better, egginger2025quantum, alvarezestevez2024benchmarking} indicating that matched-baseline benchmarking is necessary methodology for QML claims at NISQ scale. The topology-aligned framework introduced here is itself substrate-agnostic and offers a clean basis for future comparisons.

The framework also lends itself well to scaling beyond QM9. The shared parameters (atom and bond embeddings) are constant in the number of atoms; only the per-node update scales linearly. For Iso-QGNN, circuit depth scales with the maximum graph degree, which is bounded by a small constant for organic molecules. Whether this favourable scaling translates into preserved or improved performance at $N > 9$, particularly on datasets covering larger drug-like molecules, is an open question that follows naturally from the present work.

Several other open directions follow. Chemically informed initialisation of the bond and atom parameters, exploiting the same topology-alignment that drives the inductive bias, is a promising route to faster and more stable convergence than the random initialisation used here, and could provide a bridge to chemistry-aware constraint or verification schemes. Equivariant extensions that incorporate three-dimensional geometric information could address the bottleneck both architectures face on the dipole task. The matched-comparison framework itself extends straightforwardly to other inductive biases relevant in chemistry beyond bond topology. Together, these directions provide a concrete path for testing where quantum implementations of topology-aligned architectures genuinely outperform classical ones, rather than where they merely coexist with them.

\section*{Data and Code Availability}
The QM9 dataset is publicly available through DeepChem. Source code for both Iso-QGNN and Iso-CGNN, together with the matched-comparison training pipeline, the non-graph control, and the scripts reproducing every figure and table in this paper, is publicly available at \url{https://github.com/QunaSys/IsoQGNN}.

\bibliographystyle{IEEEtran}
\bibliography{IEEEabrv,references}

\end{document}